%% file: arxiv.tex
\documentclass{article}

\usepackage{amsmath}
\usepackage{amsfonts}
\usepackage{graphicx}
\usepackage{import}
\usepackage{pgf}

\usepackage{booktabs,dcolumn, subcaption}
\usepackage{multirow}

\usepackage{hyperref}
\hypersetup{
colorlinks,
citecolor=blue,
filecolor=blue,
linkcolor=blue,
urlcolor=blue
}
\usepackage{tikz}
\usetikzlibrary{fit,positioning}

\newcommand{\X}{\mathbf{U}}
\newcommand{\W}{\mathbf{V}}
\newcommand{\Y}{\mathbf{R}}
\newcommand{\x}{\mathbf{u}}
\newcommand{\w}{\mathbf{v}}

\newcommand{\R}{\mathbb{R}} 

\title{A High-Performance Implementation of Bayesian Matrix Factorization with Limited Communication}

\author{
Tom Vander Aa, Roel Wuyts, Wilfried Verachtert\\
ExaScience Life Lab at imec, Leuven, Belgium\\
\\
Xiangju Qin, Paul Blomstedt, Samuel Kaski\\
Helsinki Institute for Information Technology (HIIT),\\
Department of Computer Science,\\
Aalto University, Finland\\
\\
\emph{tom.vanderaa@imec.be}\\
}

\begin{document}

\maketitle

\begin{abstract}

    Matrix factorization is a very common machine learning technique in recommender systems.  Bayesian Matrix Factorization (BMF) algorithms would be attractive because of their ability to quantify uncertainty in their predictions and avoid over-fitting, combined with high prediction
    accuracy. However, they have not been widely used on large-scale data because of their prohibitive computational cost. 
%
    In recent work, efforts have been made to reduce the cost,
    both by improving the scalability of the BMF algorithm as well as its implementation, but so far mainly separately.
    In this paper we show that the state-of-the-art of both approaches to scalability can be combined. We combine the recent highly-scalable Posterior Propagation algorithm for BMF, which parallelizes computation of blocks of the matrix, with a distributed BMF implementation that users asynchronous communication within each block.
    We show that the combination of the two methods gives substantial improvements in the scalability of BMF on web-scale datasets, when the goal is to reduce the wall-clock time.
\end{abstract}

\section{Introduction}

Matrix Factorization (MF) is a core machine learning technique for applications of
collaborative filtering, such as recommender systems or drug discovery, where a
data matrix $\Y$ is factorized into a product of two matrices, such that $\Y
\approx \X \W^\top$. The main task in such applications is to predict
unobserved elements of a partially observed data matrix. In recommender
systems, the elements of $\Y$ are often ratings given by users to items, while
in drug discovery they typically represent bioactivities between chemical
compounds and protein targets or cell lines. 

Bayesian Matrix Factorization (BMF) \cite{bhattacharya2011sparse,Salakhutdinov+Mnih:2008b}, 
formulates the matrix factorization task as
a probabilistic model, with Bayesian inference conducted on the unknown
matrices  $\X$ and $\W$.  Advantages often associated with BMF include
robustness to over-fitting and improved predictive accuracy, as well as
flexible utilization of prior knowledge and side-data.  Finally, for
application domains such as drug discovery, the ability of the Bayesian
approach to quantify uncertainty in predictions is of crucial importance 
\cite{10.1093/bioinformatics/btz335}.

Despite the appeal and many advantages of BMF,
scaling up the posterior inference for industry-scale problems has proven
difficult. Scaling up to this level requires both data and
computations to be distributed over many workers, and so far only very few distributed
implementations of BMF have been presented in the literature. 
In \cite{VanderAa+others:2017}, a high-performance computing implementation of BMF using Gibbs 
sampling for distributed systems was proposed. The authors considered three different 
distributed programming models: Message Passing Interface (MPI),  Global Address Space 
Programming Interface (GASPI) and ExaSHARK. 
In a different line of work, \cite{Ahn+others:2015} proposed to use a distributed version 
of the minibatch-based Stochastic Gradient Langevin Dynamics algorithm for posterior inference 
in BMF models. 
%
While a key factor in devising efficient distributed solutions is to be able to minimize 
communication between worker nodes, both of the above solutions require some degree of communication
in between iterations. 

A recent promising proposal, aiming at minimizing communication and thus
reaching a solution in a faster way, is to use a hierarchical embarrassingly
parallel MCMC strategy \cite{Qin2019}. This technique, called BMF with
Posterior Propagation (BMF-PP),  enhances regular embarrassingly parallel MCMC
(e.g. \cite{Neiswanger:2014:AEE:3020751.3020816,Wang+others:2015}), which does
not work well for matrix factorization \cite{Qin2019} because of
identifiability issues. BMF-PP introduces communication at predetermined
limited phases in the algorithm to make the problem identifiable, effectively
building one model for all the parallelized data subsets, while in previous
works multiple independent solutions were found per subset.

The current paper is based on a realization that the approaches of
\cite{VanderAa+others:2017} and \cite{Qin2019} are compatible, and in fact
synergistic. BMF-PP will be able to parallelize a massive matrix but will be
the more accurate the larger the parallelized blocks are. Now replacing the
earlier serial processing of the blocks by the distributed BMF in
\cite{VanderAa+others:2017} will allow making the blocks larger up to the
scalability limit of the distributed BMF, and hence decrease the wall-clock
time by engaging more processors.

The main contributions of this work are:

\begin{itemize}
\item We combine both approaches, allowing for parallelization
both at the algorithmic and at the implementation level. 
\item We analyze what is the best way to subdivide the original matrix into
subsets, taking into account both compute performance \emph{and} model
quality.
\item We examine several web-scale datasets and show that datasets with
different properties (like the number of non-zero items per row) require
different parallelization strategies.
\end{itemize}

The rest of this paper is organized as follows. In Section~\ref{sec:algo}
we present the existing Posterior Propagation algorithm and distributed
BMF implementation and we explain how to combine both. Section~\ref{sec:exp} is
the main section of this paper, where we document the experimental setup and
used dataset, we compare with related MF methods, and present the results both
from a machine learning point of view, as from a high-performance compute point
of view. In Section~\ref{sec:conclusions} we draw conclusions and propose future
work.

\section{Distributed Bayesian Matrix Factorization with Posterior Propagation}
\label{sec:algo}

In this section, we first briefly review the BMF model and then describe the individual 
aspects of distributed computation and Posterior Propagation and how to combine them.

\subsection{Bayesian Matrix Factorization}

In matrix factorization, a (typically very sparsely observed) data matrix
$\Y\in \R^{N\times D}$ is factorized into a product of two matrices $\X \in
\R^{N\times K}= (\x_1,\ldots,\x_N)^\top$ and $\W = (\w_1,\ldots,\w_D)^\top\in
\R^{D\times K}$. In the context of recommender systems, $\Y$ is a rating
matrix, with $N$ the number of users and $D$ the number of rated items.

In Bayesian matrix factorization
\cite{bhattacharya2011sparse,Salakhutdinov+Mnih:2008b}, the data are
modelled as 

\begin{equation}\label{eq:BMF_likelihood_normal}
p(\Y|\X,\W) = \prod_{n=1}^N \prod_{d=1}^D \left[\mathcal{N}\left(r_{nd}|\x_n^\top\w_d,\tau^{-1}\right) \right]^{I_{nd}}
\end{equation}
where $I_{nd}$ denotes an indicator which equals 1 if the element $r_{nd}$ is
observed and 0 otherwise, and $\tau$ denotes the residual noise precision. The
two parameter matrices $\X$ and $\W$ are assigned Gaussian priors. 
Our goal is then to compute the joint posterior density $p(\X,\W|\Y) \propto p(\X)p(\W)p(\Y|\X,\W)$,
conditional on the observed data.
Posterior inference is typically done using Gibbs sampling, see \cite{Salakhutdinov+Mnih:2008b} for details.

\subsection{Bayesian Matrix Factorization with Posterior Propagation}

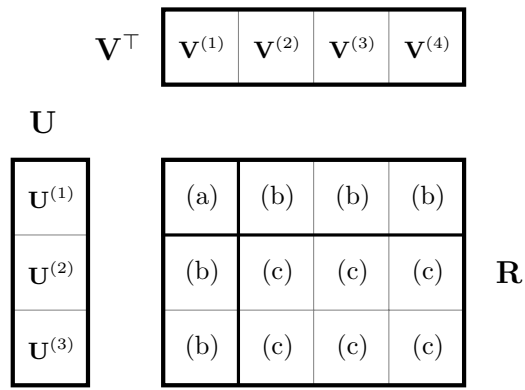
\begin{figure}
	\begin{center}
		\begin{tikzpicture}
    		\draw[help lines] (2,0) grid (6,3);
    		\node[fit={(2,0) (6,3)}, inner sep=0pt, draw=black, ultra thick] (Y) {};
    		\node[fit={(2,0) (3,3)}, inner sep=0pt, draw=black, very thick] (Yi1) {};
    		\node[fit={(2,2) (6,3)}, inner sep=0pt, draw=black, very thick] (Y1j) {};
		\node[align=center] at (2.5+0.02,2.5) {(a)};
	    \node[align=center] at (3.5+0.02,2.5) {(b)};
		\node[align=center] at (4.5+0.02,2.5) {(b)};
	    \node[align=center] at (5.5+0.02,2.5) {(b)};
		\node[align=center] at (2.5+0.02,1.5) {(b)};
	    \node[align=center] at (3.5+0.02,1.5) {(c)};
		\node[align=center] at (4.5+0.02,1.5) {(c)};
	    \node[align=center] at (5.5+0.02,1.5) {(c)};
	    	\node[align=center] at (2.5+0.02,0.5) {(b)};
	    \node[align=center] at (3.5+0.02,0.5) {(c)};
		\node[align=center] at (4.5+0.02,0.5) {(c)};
	    \node[align=center] at (5.5+0.02,0.5) {(c)};
    		\draw[help lines] (0,0) grid (1,3);
    		\node[fit={(0,0) (1,3)}, inner sep=0pt, draw=black, ultra thick] (X) {};
	    \node[align=center,font=\small] at (0.5+0.02,2.5) {$\X^{(1)}$};
	    \node[align=center,font=\small] at (0.5+0.02,1.5) {$\X^{(2)}$};
	    	\node[align=center,font=\small] at (0.5+0.02,0.5) {$\X^{(3)}$};
    		\draw[help lines] (2,4) grid (6,5);
    		\node[fit={(2,4) (6,5)}, inner sep=0pt, draw=black, ultra thick] (W) {};
	    \node[align=center,font=\small] at (2.5+0.02,4.5) {$\W^{(1)}$};
	    \node[align=center,font=\small] at (3.5+0.02,4.5) {$\W^{(2)}$};
	    	\node[align=center,font=\small] at (4.5+0.02,4.5) {$\W^{(3)}$};
	    \node[align=center,font=\small] at (5.5+0.02,4.5) {$\W^{(4)}$};
    	    \node[align=center,font=\large] at (6.5+0.1,1.5) {$\Y$};
    	    \node[align=center,font=\large] at (0.5-0.1,3.5) {$\X$};
    	    \node[align=center,font=\large] at (1.5-0.1,4.5) {$\W^\top$};
		\end{tikzpicture}
	\end{center}
	\caption{Illustration of Posterior Propagation (PP) for a data
		matrix $\Y$ partitioned into $3 \times 4$ subsets. Subset inferences for the
		matrices $\X$ and $\W$ proceed
		in three successive phases, with posteriors obtained in one phase being
		propagated as priors to the next one. The letters (a), (b), (c) in the matrix $\Y$ refer to the
		phase in which the particular data subset is processed. Within each phase, the
		subsets are processed in parallel with no communication. Figure adapted from \cite{Qin2019}.}\label{fig:PP}
\end{figure}

In the \emph{Posterior Propagation} (PP) framework \cite{Qin2019}, we start by partitioning
$\Y$ with respect to both rows and columns into $I\times J$ subsets
$\Y^{(i,j)}$, $i=1,\ldots,I$, $j = 1,\ldots, J$. 
The parameter matrices $\X$ and $\W$ are correspondingly partitioned into $I$ and $J$ submatrices, respectively. 
The basic idea of PP is to
process each subset using a hierarchical embarrassingly parallel MCMC scheme in
three phases, where the posteriors from each phase are propagated forwards and
used as priors in the following phase, thus introducing dependencies between
the subsets. The approach proceeds as follows (for an illustration, see
Figure~\ref{fig:PP}):

\paragraph{Phase (a):} 
    Joint inference for submatrices $(\X^{(1)}, \W^{(1)})$, conditional on data subset $\Y^{(1,1)}$:
    \begin{align*}
    &p\left(\X^{(1)},\W^{(1)}|\Y^{(1,1)}\right) \propto  p\left(\X^{(1)}\right)p\left(\W^{(1)}\right) p\left(\Y^{(1,1)}|\X^{(1)},\W^{(1)}\right).
    \end{align*}
\paragraph{Phase (b):} 
    Joint inference in parallel for submatrices $(\X^{(i)},\W^{(1)})$, $i=2,\ldots,I$, and $(\X^{(1)},\W^{(j)})$, $j=2,\ldots,J$, conditional on data subsets which share columns or rows with $\Y^{(1,1)}$, and using posterior marginals from phase (a) as priors:
    \begin{align*}
    p\left(\X^{(i)},\W^{(1)}|\Y^{(1,1)},\Y^{(i,1)}\right) & \propto  p\left(\W^{(1)}|\Y^{(1,1)}\right) p\left(\X^{(i)}\right) p\left(\Y^{(i,1)}|\X^{(i)},\W^{(1)}\right),\\
    p\left(\X^{(1)},\W^{(j)}|\Y^{(1,1)},\Y^{(1,j)}\right) & \propto  p\left(\X^{(1)}|\Y^{(1,1)}\right) p\left(\W^{(j)}\right) p\left(\Y^{(1,j)}|\X^{(1)},\W^{(j)}\right). 
    \end{align*}
\paragraph{Phase (c):} 
    Joint inference in parallel for submatrices $(\X^{(i)},\W^{(j)})$, $i=2,\ldots,I$, $j=2,\ldots,J$, conditional on the remaining data subsets, and using posterior marginals propagated from phase (b) as priors:
    \begin{align*}
    &p\left(\X^{(i)},\W^{(j)}|\Y^{(1,1)},\Y^{(i,1)},\Y^{(1,j)},\Y^{(i,j)}\right)  \\
    &\propto p\left(\X^{(i)}|\Y^{(1,1)},\Y^{(i,1)}\right) p\left(\W^{(j)}|\Y^{(1,1)},\Y^{(1,j)}\right) 
    p\left(\Y^{(i,j)}|\X^{(i)},\W^{(j)}\right). 
    \end{align*}    

Finally, the aggregated posterior is obtained by combining the posteriors obtained in phases (a)-(c) and  dividing away the multiply-counted propagated posterior marginals; see \cite{Qin2019} for details. 

\subsection{Distributed Bayesian Matrix Factorization}

In \cite{VanderAa+others:2017} a distributed parallel implementation of BMF is
proposed. In that paper an implementation the BMF algorithm~\cite{Salakhutdinov+Mnih:2008b} is
proposed that distributes the rows and columns of $\Y$ across
different nodes of a supercomputing system.

\begin{figure}
  \centering
  \def\svgwidth{.7\columnwidth}
  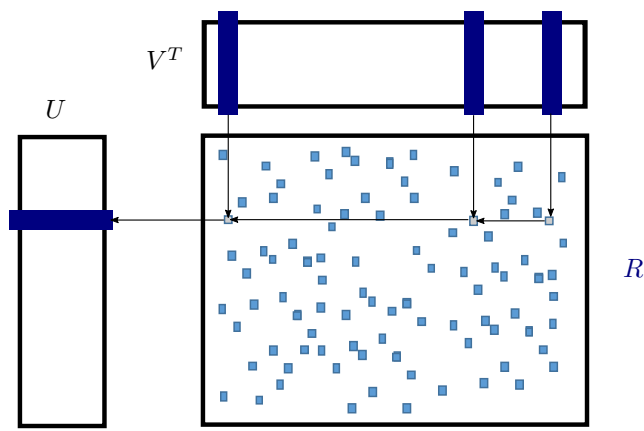
  \caption{Communication pattern in the distributed implementation of BMF}
  \label{fig:comm_struct}
\end{figure}

Since Gibbs sampling is used, rows of $\X$ and rows of $\W$ are independent
and can be sampled in parallel, on different nodes. However, there is a dependency
between samples of $\X$ and $\W$. The communication pattern
between $\X$ and $\W$ is shown in Figure~\ref{fig:comm_struct}.

The main contribution of this implementation is how to distribute $\X$ and $\W$
to make sure the computational load is distributed as equally as possible
and the amount of data communication is minimized.  The authors of
\cite{VanderAa+others:2017} optimize the distributions by analysing the sparsity structure
of $\Y$. 

As presented in the paper, distributed BMF provides a reasonable speed-up for
compute systems up to 128 nodes. After this, speed-up in the strong scaling
case is limited by the increase in communication, while at the same
computations for each node decrease.

\subsection{Distributed Bayesian Matrix Factorization with Posterior Propagation}

By combining the distributed BMF implementation with PP, we
can exploit the parallelism at both levels. The different subsets of $\Y$ in
the parallel phases of PP are independent and can thus be computed on in
parallel on different resources.  Inside each subset we exploit parallelism by
using the distributed BMF implementation with MPI communication.

%

\section{Experiments}
\label{sec:exp}

In this section we evaluate the distributed BMF implementation with Posterior
Propagation (D-BMF+PP) and competitors 
for benchmark
datasets. We first have a look at the datasets we will use for the experiments.
Next, we compare our proposed D-BMF+PP implementation with other related matrix
factorization methods in terms of accuracy and runtime, with a special
emphasis on the benefits of Bayesian methods. 
Finally we look at the strong scaling of D-BMF+PP and what is a good way to split the
matrix in to blocks for Posterior Propagation.


\subsection{Datasets}

\begin{table}
\centering
\caption{Statistics about benchmark datasets. }
    \begin{tabular}{l|r|r|r|r|r}
    Name  & Movielens \cite{Harper:movielens} & Netflix \cite{Gomez-Uribe:2015:NRS:2869770.2843948} 
& Yahoo \cite{Yahoo} & Amazon \cite{Amazon}\\
\hline
    Scale &  1-5  &  1-5  &  0-100 & 1-5 \\
    \# Rows & 138.5K & 480.2K & 1.0M  & 21.2M \\
    \# Columns & 27.3K & 17.8K & 625.0K & 9.7M \\
    \# Ratings & 20.0M & 100.5M & 262.8M & 82.5M \\
\hline
    Sparsity & 189   & 85    &  2.4K  & 2.4M \\
    Ratings/Row & 144   & 209   & 263   & 4 \\
    \#Rows/\#Cols & 5.1   & 27.0 & 1.6   & 2.2 \\
\hline
    K     & 10    & 100   & 100   & 10 \\
    rows/sec ($\times 1000$) & 416   & 15   & 27    & 911 \\
    ratings/sec ($\times 10^6$) & 70    & 5.5  & 5.2   & 3.8 \\
    \end{tabular}%
   \label{tab:datasets}%
\end{table}%

Table~\ref{tab:datasets} shows the different webscale datasets we used for
our experiments. The table shows we have a wide diversity of properties: from
the relatively small Movielens dataset with 20M ratings, to the 262M ratings
Yahoo dataset.  Sparsity is expressed as the ratio of the total number of
elements in the matrix (\#rows $\times$ \#columns) to the number of filled-in
ratings (\#ratings). This metric also varies: especially the Amazon dataset is
very sparsely filled.

The scale of the ratings (either 1 to 5 or 0 to 100) is only important when
looking at prediction error values, which we will do in the next section.

The final three lines of the table are not data properties, but rather properties
of the matrix factorization methods. K is the number of latent dimensions used, which 
is different per dataset but chosen to be common to all matrix factorization methods.  The
last two lines are two compute performance metrics, which will be discussed in
Section~\ref{sec:scaling}.


\subsection{Related Methods}

In this section we compare the proposed BMF+PP method to other matrix
factorization (MF) methods, in terms of accuracy (Root-Mean-Square Error or
RMSE on the test set), and in terms of compute performance (wall-clock time).

Stochastic gradient decent (SGD \cite{Teflioudi2012:DSGD++}), Coordinate
Gradient Descent (CGD \cite{Yu2012:ccd++}) and Alternating Least Squares (ALS
\cite{Tan2016:FCP}) are the three most-used algorithms for non-Bayesian MF.

CGD- and ALS-based algorithms update along one dimension at a time while the other
dimension of the matrix remains fixed. Many variants of ALS and CGD exist that
improve the convergence \cite{hsieh2011fastcoord}, or the parallelization degree
\cite{Pilszy2010FastAM}, or are optimized for  non-sparse rating matrices
\cite{parallelSGD}.  
In this paper we limit ourselves to comparing to methods, which divide the rating 
matrix into blocks, which is necessary to support very large matrices. 
Of the previous methods, this includes the SGD-based ones.

FPSGD~\cite{Teflioudi2012:DSGD++} is a very efficient SGD-based library for
matrix factorization on multi-cores. While FPSGD is a single-machine
implementation, which limits its capability to solve large-scale problems, it
has outperformed all other methods on a machine with up to 16 cores. NOMAD
\cite{Yun2014:NOMAD} extends the idea of block partitioning, adding the
capability to release a portion of a block to another thread before its full
completion. It performs similarly to FPSGD on a single machine, and can scale out
to a 64-node HPC cluster.

\begin{table}
\centering
\caption{RMSE of different matrix factorization methods on benchmark datasets.}
\label{tab:related_rmse}
\begin{tabular}{l|r|r|r}
\hline
 Dataset    & BMF+PP & NOMAD & FPSGD  \\ 
\hline
 Movielens  &  0.76 &  0.77  &  0.77  \\
 Netflix    &  0.90 &  0.91  &  0.92  \\
 Yahoo      & 21.79 & 21.91  & 21.78  \\
 Amazon     &  1.13 &  1.20  &  1.15  \\
\end{tabular}
\end{table}

Table~\ref{tab:related_rmse} compares the RMSE values on the test sets of the
four selected datasets.  For all methods, we used the $K$ provided in
Table~\ref{tab:datasets}. For competitor methods, we used the default
hyperparameters suggested by the authors for the different data sets.  As already
concluded in \cite{Qin2019}, BMF with Posterior Propagation results in
equally good RMSE compared to the original BMF. We also see from the table
that on average the Bayesian method produces only slightly better results,
in terms of RMSE. However we do know that Bayesian methods have significant
statistical advantages \cite{10.1093/bioinformatics/btz335} that stimulate
their use.

\begin{table}
\centering
    \caption{Wall-clock time (hh:mm) of different matrix
    factorization methods on benchmark datasets, running on
    single-node system with 16 cores.}
\label{tab:related_time}
\begin{tabular}{l|r|r|r|r}
\hline
Dataset & BMF+PP & BMF & NOMAD & FPSGD  \\ 
\hline
Movielens  & 0:07  &  0:14 & 0:08  & 0:09 \\ 
Netflix    & 2:02  &  4:39 & 0:08  & 1:04 \\
Yahoo      & 2:13  & 12:22 & 0:10  & 2:41 \\
Amazon     & 4:15  & 13:02 & 0:40  & 2:28 \\
\end{tabular}
\end{table}

For many applications, this advantage outweighs the much higher
computational cost. Indeed, as can be seem from Table~\ref{tab:related_time},
BMF is significantly more expensive, than NOMAD and FPSGD, even when taking in to
account the speed-ups thanks to using PP. NOMAD is the fastest method, thanks
to the aforementioned improvements compared to FPSGD.

\subsection{Block Size}

\begin{figure}
   \centering
   \includegraphics[width=0.9\columnwidth]{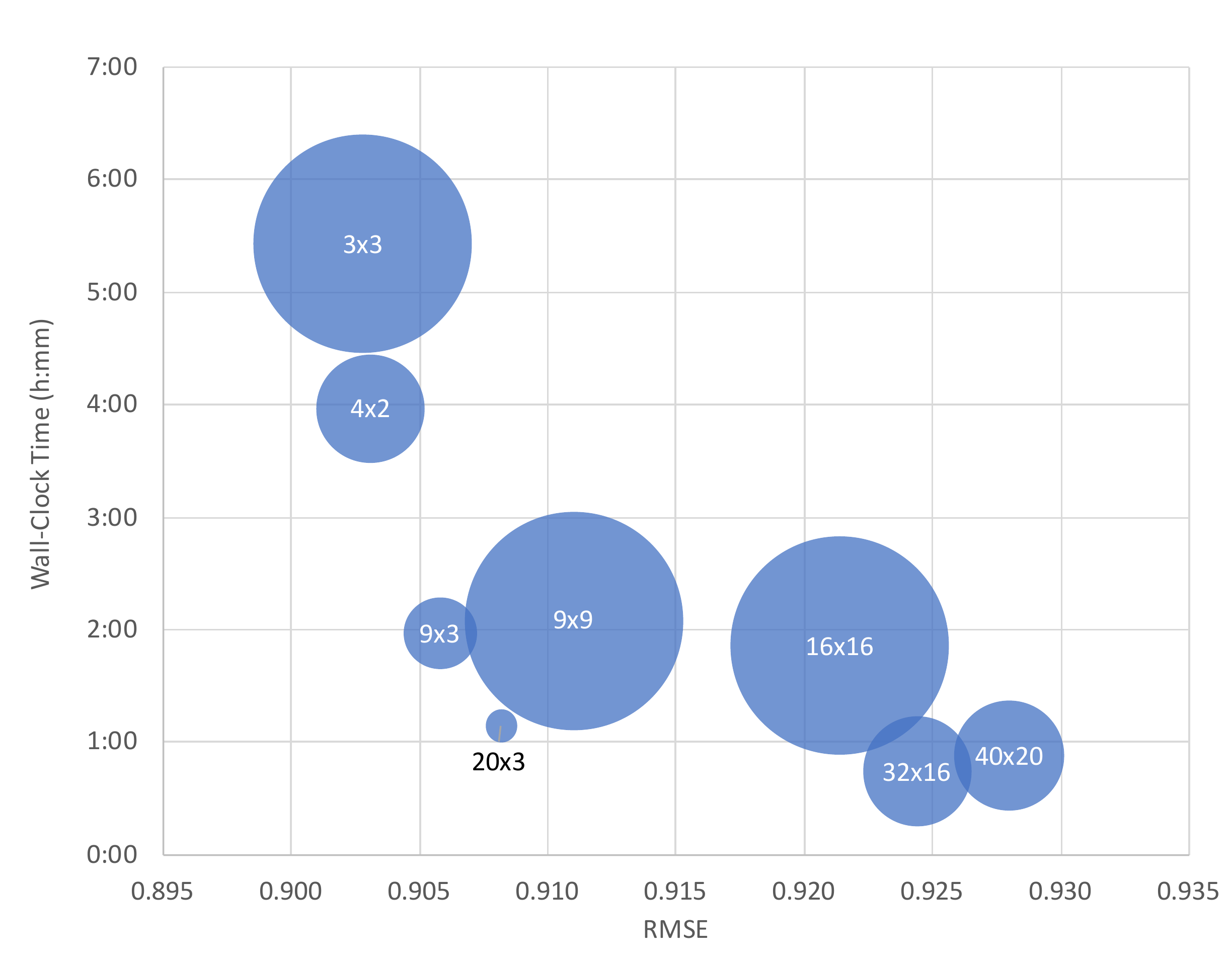}
    \caption{Block size exploration: RMSE on test set and wall-clock time (hh:mm) for different block size on Netflix.
	Each bubble is the result for a different block size indicated as the number of blocks across the rows and the columns
        inside the bubble. The size of bubbles is an indication for the aspect ratio of the
	blocks inside PP. Smaller bubbles indicate the blocks are more square.}
\label{fig:rmse_vs_blocksize}
\end{figure}

The combined method will achieve some of the parallelization with the PP
algorithm, and some with the distributed BMF within each block. We next compare
the performance as the share between the two is varied by varying the block
size.  We find that blocks should be approximately square, meaning the number
of rows and columns inside the each block should be more or less equal. This
implies the number of blocks across the rows of the $\Y$ matrix will be less if
the $\Y$ matrix has fewer rows and vice versa. 

Figure~\ref{fig:rmse_vs_blocksize} shows optimizing the block size is crucial to
achieve a good speed up with BMF+PP and avoid compromising the quality of the
model (as measured with the RMSE).  The block size in the figure, listed as
$I\times J$, means the $R$ matrix is split in $I$ blocks (with equal amount of rows)
in the vertical direction and $J$ blocks (with equal amount of columns) in the
horizontal direction.  The figure explores different block sizes for the Netflix
dataset, which has significantly more rows than columns ($27\times$) as can be seen
from Table~\ref{tab:datasets}. This is reflected by the fact that the data point
with the smallest bubble area ($20 \times 3$) provides the best trade-off between
wall-clock time and RMSE.

We stipulate that the reason is the underlying trade-off between the amount of
information in the block and the amount of compute per block. Both the amount of 
information and the amount of compute can optimized (maximized and minimized
respectively) by making the blocks approximately squared, since both are proportionate
to the ratio of the area versus the circumference of the block.

We will come back to this trade-off when we look at
performance of BMF+PP when scaling to multiple nodes for the different datasets in
Section~\ref{sec:scaling}.

\subsection{Scaling}
\label{sec:scaling}

In this section we look at how the added parallelization of Posterior
Propagation increases the strong scaling behavior of BMF+PP. Strong scaling
means we look at the speed-up obtained by increasing the amount of compute
nodes, while keeping the dataset constant.

The graphs in this section are displayed with a logarithmic scale on both axes.
This has the effect that linear scaling (i.e. doubling of the amount of
resources results in a halving of the runtime) is shown as a straight line . Additionally,
we indicate Pareto optimal solutions with a blue dot. For these solutions one
cannot reduce the execution time without increasing the resources. 

\begin{figure}
    \centering
    \includegraphics[width=0.9\columnwidth]{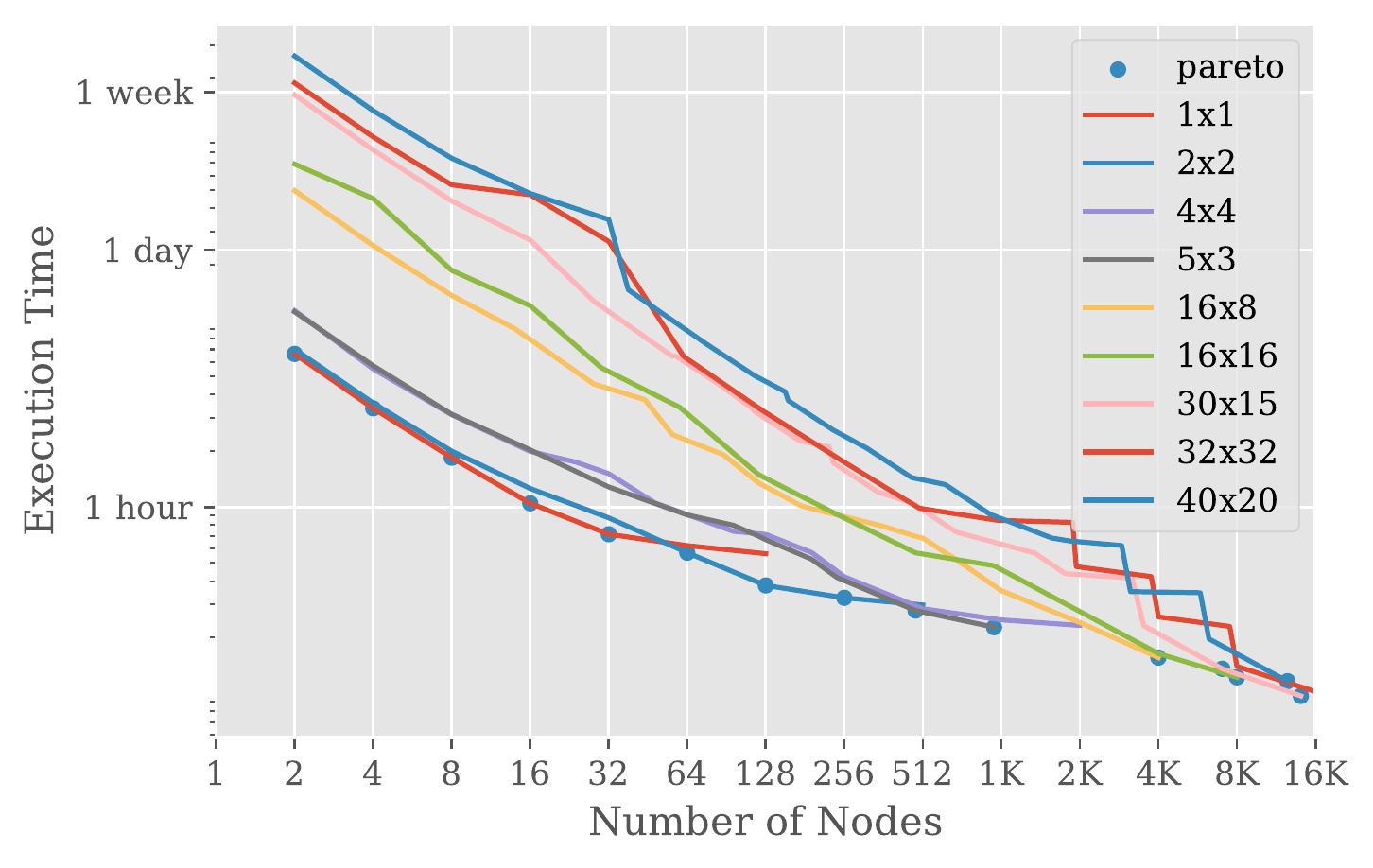}
    \includegraphics[width=0.9\columnwidth]{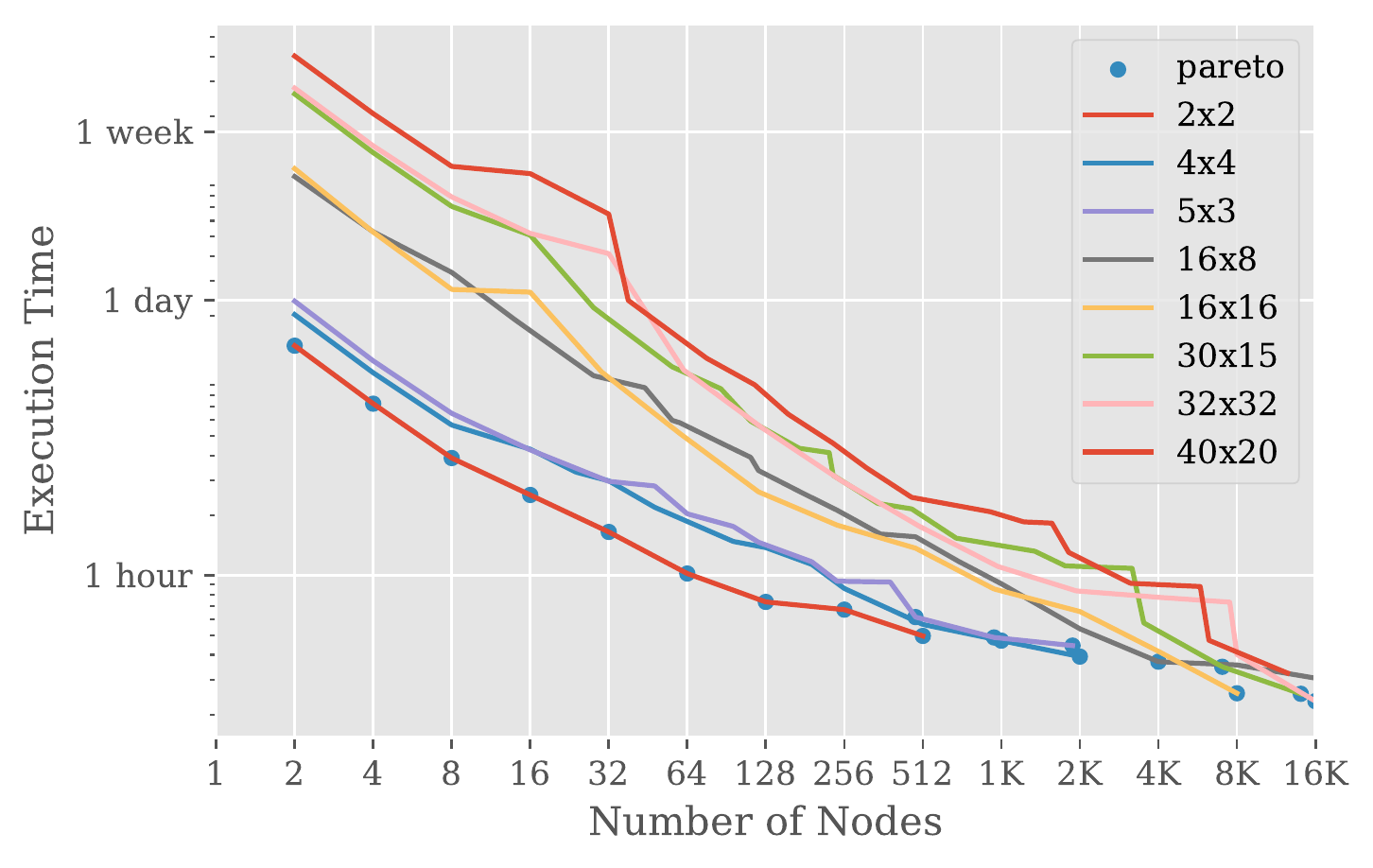}
    \caption{Strong scaling results for the Netflix (top) and Yahoo (bottom) datasets.
    X-axis indicates the amount of compute resources (\#nodes). Y-axis is wall-clock time.
    The different series correspond to different block sizes.}
    \label{fig:netflix_yahoo}
\end{figure}

We performed experiments with different block sizes, indicated with different
colors in the graphs. For example, the yellow line labeled $16 \times 8$ in Figure
Figure~\ref{fig:netflix_yahoo} means we performed PP with $16$ blocks for the rows of $\Y$
and $8$ blocks for the columns of $\Y$.

We have explored scaling on a system with up to 128K nodes, and 
use this multi-node parallelism in two ways:

\begin{enumerate}
    \item Parallelisation inside a block using the distributed version of
        BMF~\cite{VanderAa+others:2017}.

    \item Parallelism across blocks. In phase (b) we can exploit parallelism
        across the blocks in the first row and first column, using up to $I + J$ nodes
        where $I$ and $J$ is the number of blocks in the vertical, respectively horizontal
        direction of the matrix. In phase (c) we can use up to $I \times J$ nodes.
\end{enumerate}

\begin{figure}
    \centering
    \includegraphics[width=0.9\columnwidth]{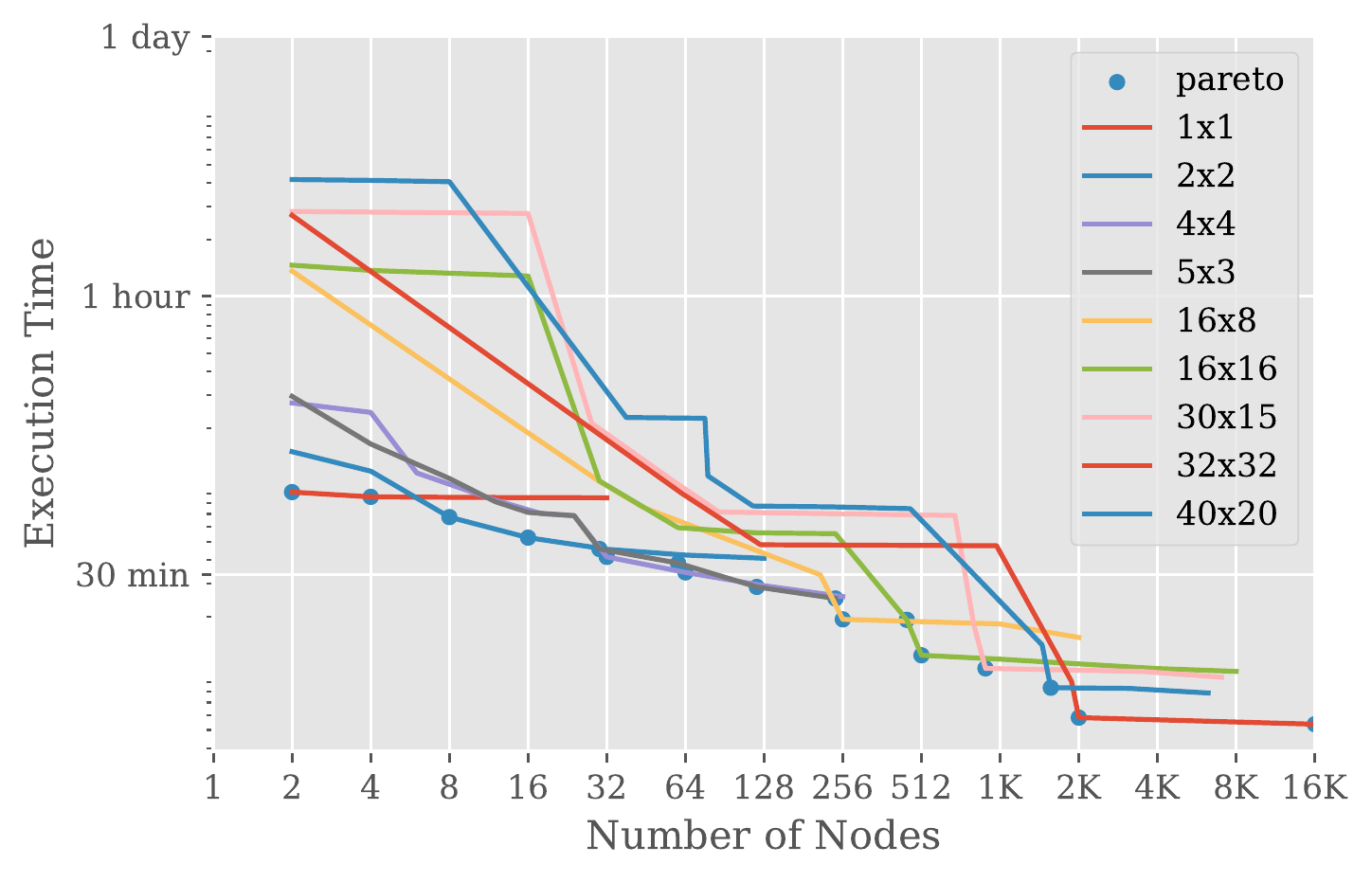}
    \includegraphics[width=0.9\columnwidth]{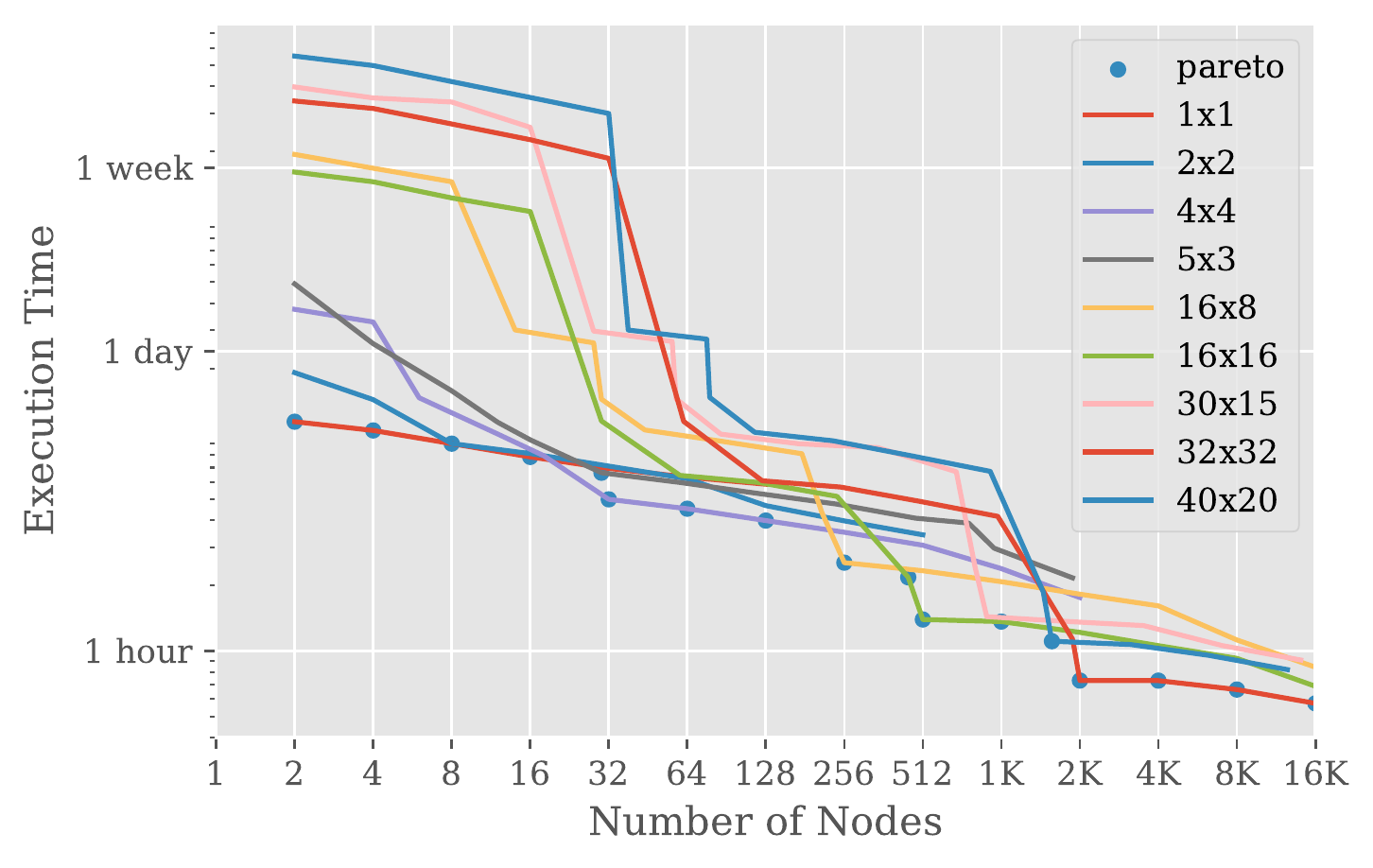}
    \caption{Strong scaling results for the Movielens (top) and Amazon (bottom) datasets.
    X-axis indicates the amount of compute resources (\#nodes). Y-axis is wall-clock time.
    The different series correspond to different block sizes.}
    \label{fig:movielens_amazon}
\end{figure}

\paragraph{General Trends}
When looking at the four graphs (Figure~\ref{fig:netflix_yahoo}
and~\ref{fig:movielens_amazon}), we observe that for the same amount of nodes
using more blocks for posterior propagation, increases the wall-clock time. The
main reason is that we do significantly more compute for this solution, because
we take the same amount of samples for each sub-block. This means for a
partitioning of $32 \times 32$ we take $1024 \times$ more samples than $1
\times 1$. 

On the other hand, we do get significant speedup, with up to $68\times$
improvement for the Netflix dataset. However the amount of resources we need
for this is clearly prohibitively large (16K nodes). To reduce execution time,
we plan to investigate reducing the amount of Gibbs samples and the effect this
has on RMSE.

\paragraph{Netflix and Yahoo}
Runs on the Netflix and Yahoo datasets (Figure~\ref{fig:netflix_yahoo})
have been performed with 100 latent dimensions (K=100).
Since computational intensity (the amount of compute per row/column of $\Y$),
is $O(K^3)$, the amount of compute versus compute for these experiments is
high, leading to good scalability for a single block ($1 \times 1$) for Netflix, and almost
linear scalability up to 16 or even 64 nodes. The Yahoo dataset is too large
to run with a $1 \times 1$ block size, but we see a similar trend for $2 \times 2$.

\paragraph{Movielens and Amazon}
For Movielens and Amazon (Figure~\ref{fig:movielens_amazon}), we used $K=10$
for our experiments. This implies a much smaller amount of compute for the same
communication cost inside a block.  Hence scaling for $1 \times 1$ is mostly
flat. We do get significant performance improvements with more but smaller
blocks where the Amazon dataset running on 2048 nodes, with $32 \times 32$
blocks is $20 \times$ faster than the best block size ($1 \times 1$) on a
single node. 

When the nodes in the experiment align with the parallelism across of blocks
(either $I + J$ or $I \times J$ as explained above), we see a significant drop
in the run time. For example for the Amazon dataset on $32 \times 32$ blocks going
from 1024 to 2048 nodes.

\section{Conclusions and Further Work}
\label{sec:conclusions}

In this paper we presented a scalable, distributed implementation of Bayesian
Probabilistic Matrix Factorization, using asynchronous communication.  We
evaluated both the machine learning and the compute performance on several
web-scale datasets.  

While we do get significant speed-ups, resource requirements for these
speed-ups are extermely large.  Hence, in future work, we plan to investigate the effect
of the following measures on execution time, resource usage and RMSE:

\begin{itemize}
    \item Reduce the number of samples for sub-blocks in phase (b) and phase (c).
    \item Use the posterior from phase (b) blocks to make predictions for phase (c) blocks.
    \item Use the GASPI implementation of \cite{VanderAa+others:2017}, instead of the MPI implementation. 
\end{itemize}

\section*{Acknowledgments}
The research leading to these results has received funding from the European
Union’s Horizon2020 research and innovation programme under the EPEEC project,
grant agreement No 801051. The work was also supported by the Academy of Finland (Flagship programme: Finnish Center for Artificial Intelligence, FCAI; grants 319264, 292334).
We acknowledge PRACE for awarding us access to Hazel Hen at GCS@HLRS, Germany.
%

\end{document}

%% file: figs/comm_struct.pdf_tex
\begingroup%
  \makeatletter%
  \providecommand\color[2][]{%
    \errmessage{(Inkscape) Color is used for the text in Inkscape, but the package 'color.sty' is not loaded}%
    \renewcommand\color[2][]{}%
  }%
  \providecommand\transparent[1]{%
    \errmessage{(Inkscape) Transparency is used (non-zero) for the text in Inkscape, but the package 'transparent.sty' is not loaded}%
    \renewcommand\transparent[1]{}%
  }%
  \providecommand\rotatebox[2]{#2}%
  \ifx\svgwidth\undefined%
    \setlength{\unitlength}{430.55635071bp}%
    \ifx\svgscale\undefined%
      \relax%
    \else%
      \setlength{\unitlength}{\unitlength * \real{\svgscale}}%
    \fi%
  \else%
    \setlength{\unitlength}{\svgwidth}%
  \fi%
  \global\let\svgwidth\undefined%
  \global\let\svgscale\undefined%
  \makeatother%
  \begin{picture}(1,0.6537193)%
    \put(0,0){\includegraphics[width=\unitlength,page=1]{comm_struct.pdf}}%
    \put(0.96233069,0.23628754){\color[rgb]{0,0,0.50196078}\makebox(0,0)[lb]{\smash{$R$}}}%
    \put(0,0){\includegraphics[width=\unitlength,page=2]{comm_struct.pdf}}%
    \put(0.0576514,0.48553193){\color[rgb]{0,0,0}\makebox(0,0)[lb]{\smash{$U$}}}%
    \put(0,0){\includegraphics[width=\unitlength,page=3]{comm_struct.pdf}}%
    \put(0.21517395,0.56269807){\color[rgb]{0,0,0}\makebox(0,0)[lb]{\smash{$V^T$}}}%
    \put(0,0){\includegraphics[width=\unitlength,page=4]{comm_struct.pdf}}%
  \end{picture}%
\endgroup%